\documentclass[sigconf]{acmart}
\usepackage{subfigure}
\usepackage{bm}
\usepackage{multirow}
\usepackage[switch]{lineno}
\usepackage{enumitem}

\usepackage{listings}

\usepackage{algorithm}
\usepackage{algpseudocode}
\usepackage{array}
\usepackage{dcolumn}
\usepackage{soul}
\usepackage{color}
\usepackage{longtable}

\usepackage{xcolor}

\definecolor{codegreen}{rgb}{0,0.6,0}
\definecolor{codegray}{rgb}{0.5,0.5,0.5}
\definecolor{codepurple}{rgb}{0.58,0,0.82}
\definecolor{backcolour}{rgb}{0.95,0.95,0.92}

\lstdefinestyle{mystyle}{
    backgroundcolor=\color{backcolour},   
    commentstyle=\color{codegreen},
    keywordstyle=\color{magenta},
    numberstyle=\tiny\color{codegray},
    stringstyle=\color{codepurple},
    basicstyle=\ttfamily\footnotesize,
    breakatwhitespace=false,         
    breaklines=true,                 
    captionpos=b,                    
    keepspaces=true,                 
    numbers=left,                    
    numbersep=5pt,                  
    showspaces=false,                
    showstringspaces=false,
    showtabs=false,                  
    tabsize=2
}

\AtBeginDocument{%
	\providecommand\BibTeX{{%
			\normalfont B\kern-0.5em{\scshape i\kern-0.25em b}\kern-0.8em\TeX}}}

\copyrightyear{2022}
\acmYear{2022}
\setcopyright{acmcopyright}\acmConference[~]{Proceedings of the 28th ACM SIGKDD Conference on Knowledge Discovery and Data Mining}{OverleafCopilot,}{2024}

\begin{document}
	
	\title[Applied Data Science Track Paper]{OverleafCopilot: Empowering Academic Writing \\ in Overleaf with Large Language Models
}

	
	\author[Haomin Wen, et al.]{Haomin Wen$^{1,2}$, Zhenjie Wei$^{1}$, Yan Lin$^{1,2,*}$,  Jiyuan Wang$^{1}$, Yuxuan Liang$^{3}$, Huaiyu Wan$^{1,2}$}
	\affiliation{%
		\textsuperscript{\rm 1}School of Computer and Information Technology, Beijing Jiaotong University, Beijing, China\\
		\textsuperscript{\rm 2}Beijing Key Laboratory of Traffic Data Analysis and Mining, Beijing, China \\
       \textsuperscript{\rm 3}Hong Kong University of Science and Technology (Guangzhou) 
        \country{China}
	}
   
	\email{{wenhaomin, 20241068, ylincs, 20271161, hywan }@bjtu.edu.cn; yuxliang@outlook.com}


\begin{abstract}
    The rapid development of Large Language Models (LLMs) has facilitated a variety of applications from different domains. In this technical report, we explore the integration of LLMs and the popular academic writing tool, Overleaf, to enhance the efficiency and quality of academic writing. To achieve the above goal, there are three challenges:  i) including seamless interaction between Overleaf and LLMs, ii) establishing reliable communication with the LLM provider, and iii) ensuring user privacy. To address these challenges, we present OverleafCopilot, the first-ever tool (i.e., a browser extension) that seamlessly integrates LLMs and Overleaf, enabling researchers to leverage the power of LLMs while writing papers. Specifically, we first propose an effective framework to bridge LLMs and Overleaf. Then, we developed PromptGenius, a website for researchers to easily find and share high-quality up-to-date prompts. Thirdly, we propose an agent command system to help researchers quickly build their customizable agents. OverleafCopilot\footnote{https://chromewebstore.google.com/detail/overleaf-copilot/eoadabdpninlhkkbhngoddfjianhlghb} has been on the Chrome Extension Store, which now serves thousands of researchers. Additionally, the code of PromptGenius is released at https://github.com/wenhaomin/ChatGPT-PromptGenius. We believe our work has the potential to revolutionize academic writing practices, empowering researchers to produce higher-quality papers in less time.
\end{abstract}
	
	

\keywords{Large Language Models; Overleaf; ChatGPT}
	
	\maketitle

\section{Introduction}

Large Language Models (LLMs) have emerged as powerful tools in various fields, including natural language processing, machine translation, and text generation. These models, such as GPT-3\cite{gpt3.5} developed by OpenAI, are capable of understanding and generating human-like text, making them invaluable for academic writing tasks. With their ability to generate coherent and contextually relevant content, LLMs have the potential to revolutionize the way we approach academic writing.

One of the most popular online academic writing tools is Overleaf\cite{overleaf}. Overleaf provides a collaborative online platform for researchers and scholars to write, edit, and publish their academic papers. It offers a range of features such as LaTeX templates, real-time collaboration, and easy integration with reference management tools. Overleaf has gained significant popularity among researchers due to its user-friendly interface and efficient collaborative capabilities.

The motivation behind this technical report is to explore the possibilities of combining LLMs and Overleaf to empower academic writing. By leveraging the capabilities of LLMs within the Overleaf platform, researchers can potentially accelerate the process of writing academic papers while significantly improving the quality of their work. This integration has the potential to enhance the efficiency and effectiveness of academic writing, allowing researchers to produce higher-quality papers in less time.

To effectively combine LLMs and Overleaf, several key problems need to be addressed. 
(i) Firstly, there is a need to develop a seamless integration between Overleaf and LLMs. This involves incorporating LLMs into the Overleaf platform as a writing assistant, allowing researchers to utilize the capabilities of LLMs while writing their papers. 
(ii) Secondly, it is essential to establish a reliable and efficient interaction with the LLM provider. This includes ensuring access to the necessary LLM resources and maintaining a smooth flow of data between Overleaf and the LLM provider's servers. 
(iii) Lastly, the issue of protecting user privacy needs to be addressed. As LLMs often require access to large amounts of data, it is crucial to implement robust privacy measures to safeguard user information and maintain confidentiality.

Targeting the aforementioned key problems, we create OverleafCopilot, which seamlessly integrates LLMs and Overleaf to contribute to the advancement of academic writing practices. Furthermore, to achieve the customization of an agent, we propose and implement an Template Directive Engine, which is essentially a set of commands used to define the behavior of an agent. Overall, the contributions of our work are summarized as follows:

\begin{itemize}[leftmargin=*]
    \item We developed OverleafCopilot, which is the first-ever tool that bridges the Large Language Models and Overleaf to empower and accelerate academic writing in Overleaf.

    \item \textbf{Efficiency}. An elaborate framework is designed to bridge the LLM provider and operations in Overleaf, thus greatly improving the efficiency of academic writing.

    \item \textbf{Collection of High-quality Prompts.} To help researchers easily find and share high-quality and up-to-date prompts for paper revising, a website named PromptGenius has been developed, the URL is https://www.promptgenius.site/.

    \item \textbf{Highly customizable}. A Template Directive Engine 
    is designed for the customization of agents, to meet the different needs of users.
\end{itemize}

\section{The Developed OverleafCopilot}

In this section, we first introduce some of the highlighted features of OverleafCopliot, and then we introduce the detailed architecture. At last, we introduce the proposed agent command engine.


OverleafCopilot is equipped with several features, including:

\textbf{(1) Easy to use.} We aim to enable more researchers to benefit by the development of LLM in academic writing. To this end, one can use OverleafCopilot with the API key from the LLM provider (e.g., OpenAI API key), or with the license provided by us. 

\textbf{(2) Comprehensive Functions.} To enlarge the capability of OverleafCopilot, we design several agents, with each one corresponding to at least one of the following features:

\begin{itemize}[leftmargin=*]
    \item Paper Polishing: (a) Speed up paper polishing with only one click; (b) Improve polishing quality: Based on ChatGPT 3.5 and GPT-4\cite{gpt4};  
    \item Grammar Check: (a) English grammar check; (b) Chinese grammar check;
    \item  Translation: high-quality translation, while maintaining the academic style;
    \item Writing Suggestions: Provides suggestions for improving the paper;
    \item Provides high-quality prompts, a prompt database (named PromptGenius\footnote{https://www.promptgenius.site/}) is created to accommodate different high-quality prompts;
    \item Returned results can be automatically written to the clipboard.
\end{itemize}

\textbf{(3) Highly Customizable}. Different users can have various requirements and preferences while using the extension. Therefore, the following customizable features are designed:
\begin{itemize}[leftmargin=*]
    \item Customize Prompts. Researchers in different fields can have different prior knowledge. Technical speaking, incorporating that specific prior knowledge into the prompts can lead to more desired outputs. To this end, OverleafCopilot allows for customizing prompts of different agents, so that researchers can quickly build a tool that suits their styles;
    
    \item Customize Shortcuts. Revising contents in a paper is usually a high-frequency operation. To further improve the efficiency of paper writing, OverleafCopilot allows users to set shortcuts for each agent;
    \item Customize Interface Layout. OverleafCopilot has different layouts that are deeply integrated into Overleaf, to be adaptable to various writing scenarios;
\end{itemize}

\textbf{(4) Privacy}. To further protect the user's privacy, OverleafCopilot does not save the content at user's request. Instead, it only forwards the user request to the LLM provider.





\section{System Architecture}

\subsection{Overview of the Extension Framework}

\begin{figure}
    \centering
    \includegraphics[width=1.0\linewidth]{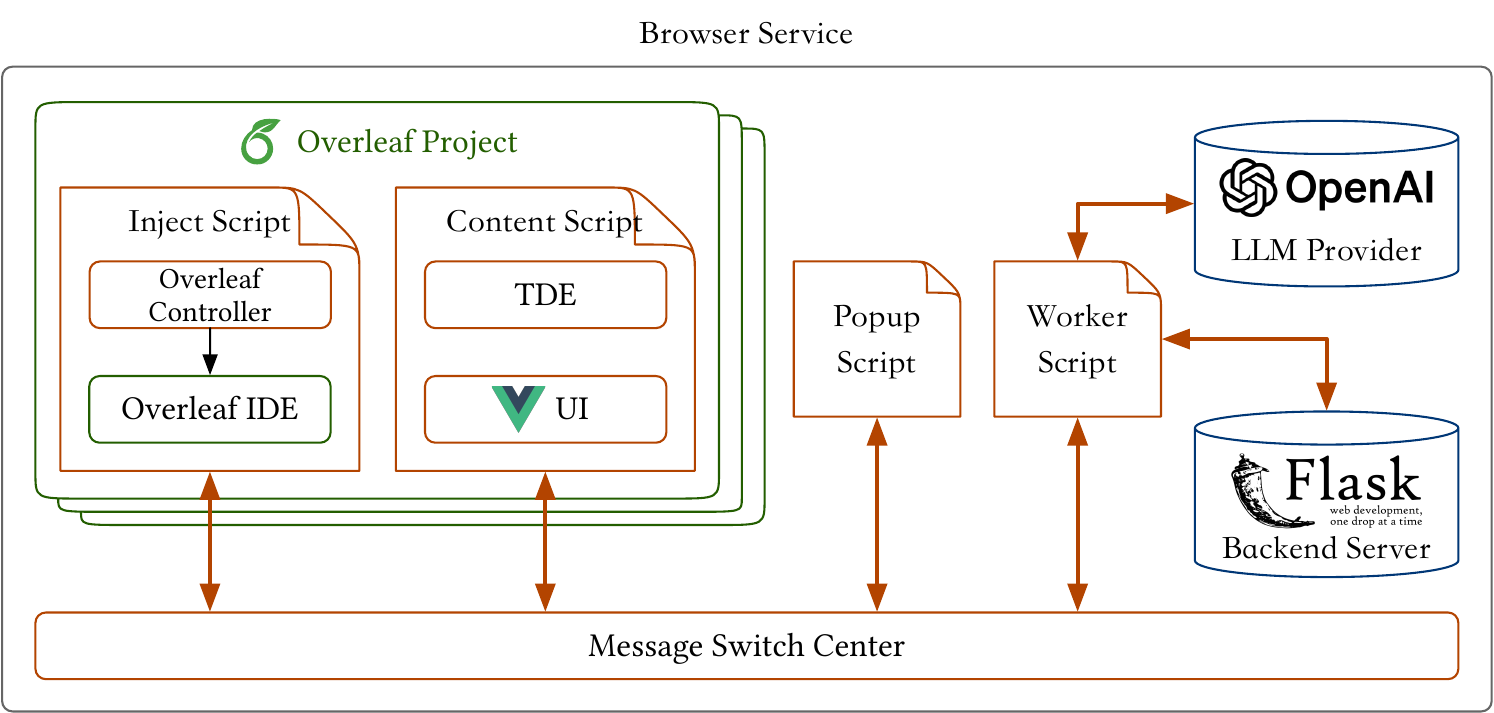}
    \caption{The overall technical framework of OverleafCopilot.}
    \label{fig:framework}
\end{figure}

\subsubsection{Basic technology stack}

Our product is built on Chrome extension technology. Chrome extensions are a development technology used to customize and enhance the functionality of the Google Chrome browser. Additionally, we leverage some JavaScript frameworks and packages including Vue\cite{vue} and Vuetify to accelerate our development progress.

Vue is a modern JavaScript framework for building user interfaces. It adopts a component-based development approach, allowing us to divide their applications into reusable components, thereby improving code maintainability and scalability. Vue also provides reactive data binding and virtual DOM management, enabling us to efficiently manage and update the state of their applications. 
Vuetify, on the other hand, is a UI framework based on Vue. Following the design guidelines of Material Design, it offers a rich set of pre-defined components such as buttons, cards, forms, etc., along with various styling and theming options. 
By utilizing Vue and Vuetify, we can directly utilize the well-designed components and styles right out of the box to save time, rather than design and implement UI components from scratch.

\subsubsection{Multi-module collaborative architecture}

Based on numerous excellent open-source frameworks, we construct our own system architecture. 

Firstly, we referred to the publish-subscribe design pattern to design an event bus system, which establishes a loose coupling relationship between publishers and subscribers through a centralized event system. Innovatively, we introduce dots in the event string to hierarchically divide the scope of events. We refer to this system as the Scoped Event Bus System (or SEB for short). It serves as the cornerstone of our autonomously designed system architecture, facilitating collaboration among various modules.

On top of SEB, we develop a Message Switch Center (MSC for short) for passing messages between different scripts in a Chrome extension. In our architecture, there are four scripts that work together to achieve the overall functionality of the plugin, including the \textit{content script}, \textit{worker script}, \textit{injected script}, and \textit{popup script}, as shown in Figure~\ref{fig:framework}.


For example, when a user wishes to send the input content to OpenAI and write the obtained result into the Overleaf editor, we need to obtain the user input in the content script, notify the background worker through MSC to initiate a request to the OpenAI model, and then instruct the injected script through MSC to call the Overleaf API to render the content.

\subsubsection{Agents and templates}

In the field of AI, the concept of an agent is frequently mentioned. An agent refers to an entity that is capable of perceiving the environment and making decisions and actions in a certain way. It can interact with the environment based on predefined goals and tasks. The working principle of an agent typically follows a Perceive-Think-Act cycle. Firstly, the agent perceives information from the environment through sensors such as vision, sound, and touch. Then, the agent analyzes, reasons, and makes decisions based on the perceived information using internal intelligent algorithms and knowledge base, in order to determine the next action. Finally, the agent executes the corresponding action through effectors, such as moving or manipulating objects. 

Agents are a serie of virtual entity that are user-defined and capable of perceiving user inputs and system states. They utilize AI models to process information and convert model outputs into concrete actions. When an agent is working, it goes through four basic step: executing pre-action, generate prompt, call model API and get a result, and execute post-action. These steps can be customized by the user. In the core of an agent, we have currently implemented access and usage of OpenAI GPT series model APIs, and we plan to introduce models from other providers in the future to offer users more choices and higher-quality services.

In the developed plugin, we innovatively introduce templates to define agents. A template is a tree-like structure that defines all the data and behaviors of an agent. This structure follows the XML syntax specification\cite{xml}, where each XML tag element is referred to as a directive. An XML tag consists of a tag name, attributes, and tag body. Similarly, a template directive consists of a directive name, directive parameters, and directive body. We have designed a multitude of carefully crafted directives that provide high flexibility in defining UI representations, guiding model actions, and interacting with Overleaf. By arranging and combining directives, users can freely create agents that meet their specific goals and requirements. We designed a dedicated subsystem called the Template Directive Engine (TDE) that parses and executes the directive templates, Subsequently.

\subsection{Scoped Event Bus}
\label{seb}
\subsubsection{Publish-subscribe design pattern}

\begin{figure}
    \centering
    \includegraphics[width=1\linewidth]{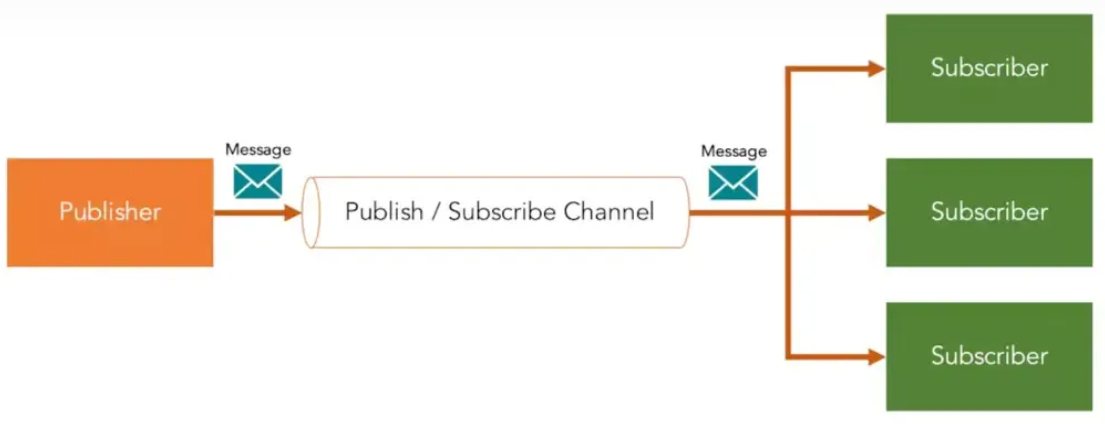}
    \caption{Publish-subscribe pattern}
    \label{fig:pub-sub}
\end{figure}

The Publish-Subscribe design pattern is a software design pattern used to achieve loosely coupled communication among components. Within this pattern, the publisher maintains a list of subscribers and notifies all subscribers when a specific event occurs. This approach allows for independence between the publisher and subscribers, enabling dynamic addition or removal of subscribers, thus facilitating flexible message transmission and decoupling.

When implementing the Publish-Subscribe design pattern, an event bus is often used as an intermediary. The event bus serves as a central dispatcher, with the publisher publishing events to the event bus, which then delivers the events to the relevant subscribers based on their registration information. The event bus can be synchronous or asynchronous, depending on the actual requirements. A synchronous event bus immediately delivers the event to subscribers, while an asynchronous event bus places the event in a queue and delivers it to subscribers at an appropriate time.

\subsubsection{Hierarchically scoped event design}

Generally speaking, events are represented by strings. When subscribing to an event, subscribers need to provide the target event string and a callback function. On the other hand, when publishing an event, the publisher needs to provide the event string and function parameters. In the developed plugin, we introduce the concept of event scopes and use dots to divide the scope within the event string.

For example, with the event string "layout. switch" published by the publisher, the following registered events will be triggered in sequence: $\langle$" (empty event), "layout", "layout. switch", "layout. switch. finally", "layout. finally", "finally"$\rangle$.

This design provides convenience for the implementation of other higher-level modules. One typical application of this design will be discussed in detail in the section on MSC.

\subsubsection{Shortcut system based on bus}

Based on the event bus system, we implement a dynamic shortcut key binding and monitoring system. When a key is pressed on the keyboard, the global event bus publishes an event string according to certain rules. For example, when the user presses the A key, the event "window.keydown.a" is published to the global event bus. As for combination keys, the listener combines the key values in a specific order to publish the corresponding event. For example, when the user presses "Control+Shift+B," the event "window.keydown.control.shift.b" is published to the global event bus. Therefore, users can register and monitor shortcut keys by listening to events on the event bus.










\subsection{Online Backend}
To support network-related operations of OverleafCopilot, such as license activation, OpenAI API Key creation, agent distribution, and timely notifications to users, we have developed an always-online backend. This backend provides network interfaces for the OverleafCopilot extension.

The backend is implemented using the Flask\footnote{https://flask.palletsprojects.com/en/3.0.x/}, which is a microweb framework written in Python. For this project, we primarily make use of its routing function to create several URL-accessible interfaces, each serving a specific function. Additionally, we leverage the wide variety of Python packages available to integrate features such as asymmetric encryption, database management, and email functionality into the backend.

The backend interfaces can be categorized into three groups: information-related, trial-related, and license-related. Information-related interfaces deliver timely notifications to users and enable the plugin to retrieve the latest version code. Trial-related interfaces enable users to initiate a new trial. License-related interfaces allow users to activate or validate a license.

\section{Conclusion}
We introduce OverleafCopilot, an innovative browser extension that seamlessly integrates LLMs (Language Model Models) with Overleaf, revolutionizing the way researchers approach academic writing.
OverleafCopilot is built upon a robust framework that effectively bridges the gap between online LLM services and Overleaf. The incorporation of built-in prompts is made possible through PromptGenius, a website developed by our team, offering researchers access to high-quality and up-to-date prompts. Furthermore, we have implemented a versatile agent command system, empowering researchers to swiftly create fully customized agents.
Currently available on the Chrome Extension Store, OverleafCopilot has already gained substantial traction, being utilized by thousands of researchers. It continues to evolve as an actively developed platform, with exciting new features planned for future releases.

{\small
\bibliographystyle{ieeetr}
\bibliography{my_bib}}

\appendix

\section{Appendix}
\subsection{Currently Supported Directives}

\subsubsection{Classification of Directive Set Groups}
\label{directive}
The supported directives in TDE are divided into different directive sets and directive set groups, each with its predefined priority and privilege level. TDE allows for the definition of directives with the same name. The priority of directive sets determines the order in which TDE searches for directive implementations and directives from directive sets with higher priority will be executed for the same directive name. TDE also assigns directive privilege levels for different target users, where users with lower privilege levels cannot access directive sets or directive set groups with higher privilege levels. During the development and beta testing phases, we do not differentiate users' privilege levels, and all users have access to directives of any privilege level. TDE also allows users to introduce custom directives using the `<predef>` directive.

In the developed plugin, we implement the following four categories of directive set groups, which meet the diverse needs of users:

\textbf{Basic Directive Set Group.} It has the lowest privilege level but higher priority. It includes the most fundamental directive sets, such as agent creation, model specification, prompt design, pre-action and post-action definition, model input/output retrieval, and basic model access directives.

\textbf{UI Control Directive Set Group.} It has a relatively lower privilege level and lower priority. It allows users to customize the UI representation of a specific agent, such as custom UI components, layouts, buttons, event bindings, and keyboard shortcut bindings. This helps improve the efficiency of user-agent interaction.

\textbf{Overleaf Interaction Directive Set Group.} It has a higher privilege level and higher priority. It allows agents to interact with the Overleaf API, such as inserting candidate content, comments, and annotations in the editor, retrieving user-selected content, and opening file contents. This enables users to customize a more automated and intelligent assistant specifically for Overleaf.

\textbf{Model Access Directive Set Group.} It has the highest privilege level but the lowest priority. In addition to the GPT Chat interface access provided by the basic directive set group, it offers access to higher-level model interfaces, such as GPT function calls, history retrieval, context access, and more. We also plan to support multimodal models and access to models provided by other providers, such as Google, in the future. This allows users to leverage the capabilities of various models to enhance their work efficiency.

\subsubsection{Directives for Defining A Typical Agent}

\begin{itemize}[leftmargin=*]
   \item  name: A short name for the agent;     
    \item icon: the agent's icon from the Material Design Icons(MDI)\footnote{https://pictogrammers.com/library/mdi/} website. 
    \item desc: the detailed description of an agent's function;
    \item model: the LLM used for the agent;
\end{itemize}

\begin{figure}
    \centering
    \includegraphics[width=1\linewidth]{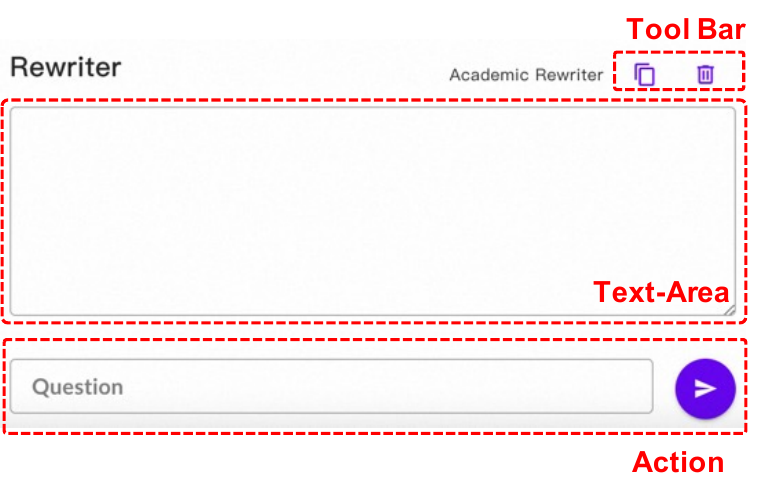}
    \caption{Illusratioin and classification of an agent workspace.}
    \label{fig:agent}
\end{figure}

\textbf{Workspace.} The second type, named <preset> describes the preset workspace and shortcut of the agent. As shown in Figure~\ref{fig:agent}, for the workspace, an agent is composed of three components, namely the toolbar, the text area, and the action. 

The text area is mainly designed to display the returned text. It has two commands including <textarea> (e.g.,  in rewriter) and <chatlist> (e.g., in translator). Moreover, if one needs to edit the text area, the command <inputarea> can be added right after the <textarea>.

\subsubsection{Other Supported Directives}

\begin{lstlisting}[language=html]  
<agent name="Rewriter">
  <icon mdi="text-box-edit-outline" />
  <desc>Academic Rewriter</desc>
  <model temperature="0.7">gpt-3.5-turbo</model>
  <preset>
    <workspace>
      <toolbar>
        <actions>
          <action preset="copy" />
          <action preset="clear" />
        </actions>
      </toolbar>
      <textarea />
      <inputarea />
      <actions>
        <action preset="send-input" bind="btn.send" />
      </actions>
    </workspace>
  </preset>
  <prompt>
    <system>I hope you act like a professional academic rewriter. I want you to revise the given content. Here are some requirements:
			1. The revised content must use the same language as the input. For example, if the input is Chinese then return the Chinese, if the input is English, then return English.
			2. Provide a revised version that maintains the original intent while improving the overall flow, clarity, and language used.
			3. Please do not utilize over-complicated words, and make changes when it is necessary.
			4. Only return the revised content.
			Here is the To-be-revised content:</system>
    <user>
      <input />
    </user>
  </prompt>
  <post-action>
    <copy />
  </post-action>
</agent>
\end{lstlisting}

\begin{lstlisting}[language=html]  
<agent name="Translator">
  <icon mdi="text-recognition" />
  <desc>Translate your paragraph</desc>
  <model temperature="0.7">gpt-3.5-turbo</model>
  <preset>
    <workspace>
      <toolbar>
        <actions>
          <action preset="copy" />
          <action preset="clear" />
        </actions>
      </toolbar>
      <chatlist />
      <inputarea />
      <actions>
        <action preset="send-input" bind="btn.send" />
      </actions>
    </workspace>
  </preset>
  <prompt>
    <system>I hope you act like a professional academic translator. I want you to translate the given content into Chinese or English based on the input language. Pay attention to the grammar, sentence structure, word choice, and clarity to enhance the readability and expression of translated content.
			Here is the input:</system>
    <user>
      <input />
    </user>
  </prompt>
  <post-action>
    <copy />
  </post-action>
</agent>

\end{lstlisting}

\begin{lstlisting}[language=html]  
<agent name="Adviser">
  <icon mdi="text-box-search-outline" />
  <desc>Academic Adviser</desc>
  <model temperature="0.7">gpt-3.5-turbo</model>
  <preset>
    <workspace>
      <toolbar>
        <actions>
          <action preset="clear" />
          <action preset="copy" />
        </actions>
      </toolbar>
      <textarea />
      <actions>
        <action preset="send-input" bind="btn.send" />
      </actions>
    </workspace>
  </preset>
  <prompt>
    <system>I hope you act like a professional academic adviser. I want you to provide some advice on the given content. To improve the overall flow, and clarity in terms of the content's language. Here is the input:</system>
    <user>
      <input />
    </user>
  </prompt>
</agent>
\end{lstlisting}

\begin{lstlisting}[language=html]  
<agent name="Checker">
  <icon mdi="text-search-variant" />
  <desc>Grammar Checker</desc>
  <model temperature="0.7">gpt-3.5-turbo</model>
  <preset>
    <workspace>
      <toolbar>
        <actions>
          <action preset="clear" />
          <action preset="copy" />
        </actions>
      </toolbar>
      <textarea />
      <inputarea />
    </workspace>
  </preset>
  <prompt>
    <system>You are an expert grammar checker in English that looks for grammar mistakes. You take all my input (maybe in latex form) and auto-correct it. Here are a few requirements:
		1. First reply to my input with the correct grammar. (Ignore the space error in the input.)
		2. Then, list the detected mistakes with each one in the following format:
		   [mistake] -&gt; [corrected content]
		3. If my input is grammatically correct and fluent, just reply ``Sounds good''.</system>
    <user>
      <input />
    </user>
  </prompt>
</agent>
\end{lstlisting}

\onecolumn
  
\begin{longtable}{ccp{0.55\textwidth}c}
\caption{This table shows the current commonly used commands and their explanations and parameters. We follow the \ref{directive} and further segment the Basic Directive Set Group into 4 small groups. All the single-row commands (i.e. func) have no parameters and the double-row commands (i.e. agent) show their explanations at the first row and parameters at the second one. The parameters shown in the table follow this format: if it's a required one, will be wrapped by '[]', and if not,'<>'. Parameters' type will be shown after ':', and if it only has limited options, we list it with '|'. Also, there are some default selections, we use '=' to show it. The parameters' explanation are wrapped by '()'.}\\
\toprule
\textbf{Class}   & \textbf{Command} & \textbf{Explanations/parameter}&  \textbf{Dependence}  \\ 
\midrule
\multirow{8}{*}{Basic.Utils} & \multirow{3}{*}{join-diff} & \texttt{Format the calculation result of the diff directive and output it as a string} & \multirow{2}{*}{-}\\
                                                            \cmidrule{3-3}
                                                        & & \texttt{[format:"html"|"markdown"|"latex"="html"](reference format)} \\
                              \cmidrule{2-4}
                            & \multirow{2}{*}{diff}       & \texttt{Calculate the difference between two texts and output as a list} & \multirow{2}{*}{-} \\
                                                             \cmidrule{3-3}                                    
                                                        & & \texttt{[unit:"line"|"word"|"letter"="word"](comparison unit)} \\                                                                                                           
                              \cmidrule{2-4}
                            & source                       & \texttt{Anchor text for diff command} & diff\\
                              \cmidrule{2-4}
                            & target                       & \texttt{The text to be compared for the diff command} & diff\\
\midrule
\multirow{13.5}{*}{Basic.Agent}  & prompt                  & \texttt{Calculate the prompt to be sent to LLM} & -\\
                              \cmidrule{2-4}
                            & system                     & \texttt{Caucualte the prompts system message(predefined)} & prompt\\  
                             \cmidrule{2-4}
                            & user                       & \texttt{Caucualte the prompts user message} & prompt\\  
                            \cmidrule{2-4}
                            & pre-action                 & \texttt{Agent action before submission} & agent\\  
                            \cmidrule{2-4}
                            & post-action                & \texttt{Agent action after submission} & agent\\  
                            \cmidrule{2-4}
                            &desc                       & \texttt{The detailed description of the agent} & -\\  
                            \cmidrule{2-4}
                             & \multirow{2}{*}{agent} & \texttt{Define an agent} & \multirow{2}{*}{-}\\
                                                            \cmidrule{3-3}
                                                        & & \texttt{[name:string](the name for the current agent)} \\
                            \cmidrule{2-4}                            
                             & \multirow{2}{*}{icon} & \texttt{The agent’s icon from the Material Design Icons(MDI) website.} & \multirow{2}{*}{-}\\
                                                            \cmidrule{3-3}
                                                        & & \texttt{[mdi:string](material design icon id)} \\
\midrule
\multirow{12}{*}{Basic.Basic}  & func                     & \texttt{Execute all commands in the command body, ignore text} & -\\
                              \cmidrule{2-4}
                            & text                       & \texttt{Splice all directive results in the directive body and return them as text} & -\\  
                             \cmidrule{2-4}
                            & copy                       & \texttt{Automatically copy the concatenation of all directive results in the directive body} & -\\  
                            \cmidrule{2-4}
                            & submit                      & \texttt{Submit prompt to LLM} & -\\  
                            \cmidrule{2-4}
                             & \multirow{2}{*}{fire}     & \texttt{Trigger an event} & \multirow{2}{*}{-}\\
                                                            \cmidrule{3-3}
                                                        & & \texttt{<event:string>(predefined event, refer to \ref{seb}) } \\
                            \cmidrule{2-4}                            
                             & \multirow{3}{*}{select*} & \texttt{Select contextual text, concatenate into a string, and return} & \multirow{3}{*}{-}\\
                                                            \cmidrule{3-3}
                                                        & & \texttt{[unit: "paragraph"|"section"|"sentence"|"word"="sentence"](offset unit), [offset:int=0] (offset,  0 represents the current unit)} \\
\midrule
\multirow{11}{*}{Basic.Buffer}  & input                    & \texttt{Get user input} & -\\
                              \cmidrule{2-4}
                            & output                       & \texttt{Obtain LLM return result} & -\\  
                             \cmidrule{2-4}
                            & \multirow{2}{*}{set-message} & \texttt{Splice the text content inside the directive body and send it to the front end} & \multirow{2}{*}{-}\\  
                            \cmidrule{2-4}
                             & \multirow{3}{*}{store} & \texttt{Store the content of the directive body (text calculation result) in the current agent's memory} & \multirow{3}{*}{-}\\
                                                            \cmidrule{3-3}
                                                        & & \texttt{<addr:string>(memory's key, memory is a dictionary)} \\
                            \cmidrule{2-4}                            
                             & \multirow{2}{*}{load} & \texttt{Load the content from the current agent's memory} & \multirow{2}{*}{-}\\
                                                            \cmidrule{3-3}
                                                        & & \texttt{<addr:string>(The key of the content be loaded)} \\
\midrule        
\multirow{3}{*}{Model Access} & \multirow{3.5}{*}{model} & \texttt{The LLM used by the agent} & \multirow{3.5}{*}{agent}\\
                                                            \cmidrule{3-3}
                                                        & & \texttt{[temperature:float=0.7](set the precision rate of the model), [max-tokens:int=2000](the maximum tokens that the model can send and receive)} \\                           
\midrule   
\multirow{20}{*}{UI Control}  & workspace                 & \texttt{Define workspace content(refer to \textbf{any} predefined agent)} & -\\
                              \cmidrule{2-4}
                            & inputarea                    & \texttt{Insert Input Box into the current agent workspace(refer to \textbf{Translator} agent)} & workspace\\  
                             \cmidrule{2-4}
                            & chatlist    & \texttt{Insert Chat List(refer to \textbf{Translator} agent)} & workspace\\  
                            \cmidrule{2-4}
                            & toolbar                      & \texttt{Insert toolbar(refer to \textbf{any} predefined agent)} & workspace\\  
                            \cmidrule{2-4}
                            & actions                      & \texttt{Insert Button Set} & -\\  
                            \cmidrule{2-4}
                            & bindings                     & \texttt{Define binding} & preset\\  
                            \cmidrule{2-4}
                             & \multirow{3}{*}{action}     & \texttt{Insert operation (UI is a button)} & \multirow{3}{*}{actions}\\
                                                            \cmidrule{3-3}
                                                        & & \texttt{[icon] (button icon)[desc] (function description) [Func] (execution function) } \\
                            \cmidrule{2-4}                            
                             & \multirow{4}{*}{richarea} & \texttt{Insert a rich text area(refer to \textbf{Checker} agent)} & \multirow{4}{*}{workspace}\\
                                                            \cmidrule{3-3}
                                                        & & \texttt{[mode:"render"|"editor"="render"](whether to enable markdown rendering mode), [switch:string](trigger event for mode switching)} \\
                            \cmidrule{2-4} 
                            & \multirow{6}{*}{keydown}     & \texttt{Define key actions (using shortcut keys,refer to \textbf{Rewriter} agent)} & \multirow{6}{*}{bindings}\\
                                                            \cmidrule{3-3}
                                                        & & \texttt{[scope:"global"|"window"|"content"="window"](shortcut key effective range), <key: string>(shortcut key setting, key combination use string representation), [present:bool=false](if set true, shortcut only take effect when the agent is in the foreground) } \\
\midrule 
\multirow{2}{*}{Overleaf Interaction}  & completion      & \texttt{Continue writing according to the context} & -\\
                              \cmidrule{2-4}
                            & insert-comment              & \texttt{Create a new comment in the review and write the result into it} & -\\  

\bottomrule
\end{longtable}

\twocolumn

\end{document}